\documentclass[11pt]{article}

\usepackage[preprint]{acl}

\usepackage{times}
\usepackage{latexsym}

\usepackage[T1]{fontenc}

\usepackage[utf8]{inputenc}

\usepackage{microtype}

\usepackage{inconsolata}

\usepackage{graphicx}

\usepackage{algorithm}
\usepackage{algorithmic}
\usepackage{amsmath}
\usepackage{adjustbox}
\usepackage{xspace}
\usepackage{amsfonts}
\usepackage{booktabs}
\usepackage{multirow}
\usepackage{xcolor}
\makeatletter 
\DeclareRobustCommand\onedot{\futurelet\@let@token\@onedot} 
\def\@onedot{\ifx\@let@token.\else.\null\fi\xspace}
\def\eg{\emph{e.g}\onedot} 
 
\def\ie{\emph{i.e}\onedot}

\usepackage[T1]{fontenc}
\newcommand\blfootnote[1]{%
  \begingroup
  \renewcommand\thefootnote{}\footnote{#1}%
  \addtocounter{footnote}{-1}%
  \endgroup
}

\title{ReContraster: Making Your Posters Stand Out with Regional Contrast}

\author{
    Peixuan Zhang$^{\#1}$, Zijian Jia$^{\#1}$, Ziqi Cai$^{2,3}$, Shuchen Weng$^{*2,4}$, Si Li$^{*1}$, Boxin Shi$^{2,3}$ \\
    \small
    $^1$School of Artificial Intelligence, Beijing University of Posts and Telecommunications \\
    \small
    $^2$State Key Laboratory for Multimedia Information Processing, School of Computer Science, Peking University\\
    \small
    $^3$National Engineering Research Center of Visual Technology, School of Computer Science, Peking University\\
    \small
    $^4$Beijing Academy of Artificial Intelligence\\
    \footnotesize
    \texttt{\{pxzhang, jiazijian, lisi\}@bupt.edu.cn}, \texttt{zqtsai@gmail.com}, \texttt{\{shuchenweng, shiboxin\}@pku.edu.cn}
}

\begin{document}

\maketitle

\begin{abstract} 
\blfootnote{$^\#$ Equal contributions. * Corresponding authors.}
Effective poster design requires rapidly capturing attention and clearly conveying messages. 
Inspired by the ``contrast effects'' principle, we propose ReContraster, the first training-free model to leverage regional contrast to make posters stand out. By emulating the cognitive behaviors of a poster designer, ReContraster introduces the compositional multi-agent system to identify elements, organize layout, and evaluate generated poster candidates. To further ensure harmonious transitions across region boundaries, ReContraster integrates the hybrid denoising strategy during the diffusion process. We additionally contribute a new benchmark dataset for comprehensive evaluation. Seven quantitative metrics and four user studies confirm its superiority over relevant state-of-the-art methods, producing visually striking and aesthetically appealing posters.

\end{abstract}

\section{Introduction}

\vspace{2mm} 
{ \textit{``It is in the contrast of light and dark that design happens.''}}
\begin{flushright}
\vspace{-1mm} 
{\textit{--Helen Van Wyk}}
\vspace{-1mm}
\end{flushright}

Posters serve as a common medium for achieving specific communicative goals~\cite{macintosh2007poster} (\eg, advertising, event promotion, and public campaigns). Unlike art forms that encourage prolonged reflection (\eg, sculptures, poetry, or documentaries), posters must convey their message and capture attention almost instantly~\cite{utoyo2021visual}. Recent text-to-poster models made significant strides in producing aesthetically coherent layouts~\cite{opencole, layoutprompter, posterllava} and detailed textual elements~\cite{textdiffuser2, anytext, glyphdraw2}. 
However, significant challenges remain in generating posters that are both visually compelling and communicatively effective from a user’s perspective.

The principle of ``contrast effects''~\cite{palmer2014theory, scherer2009contrast, o2015colour} suggests that the brain responds strongly to contrasting stimuli, quickly processing significant differences between objects and triggering heightened neural activation. David Ambarzumjan’s artwork Recover (from the ``Brushstrokes in Time'' series) exemplifies this principle, where the contrast between dark cityscapes and vibrant nature invites viewers to explore the interplay of the depicted elements\footnote{We present representative artworks in Supp.}, amplifying both visual impact and thematic clarity.

\begin{figure*}[t]
  \centering
  \includegraphics[width=\linewidth]{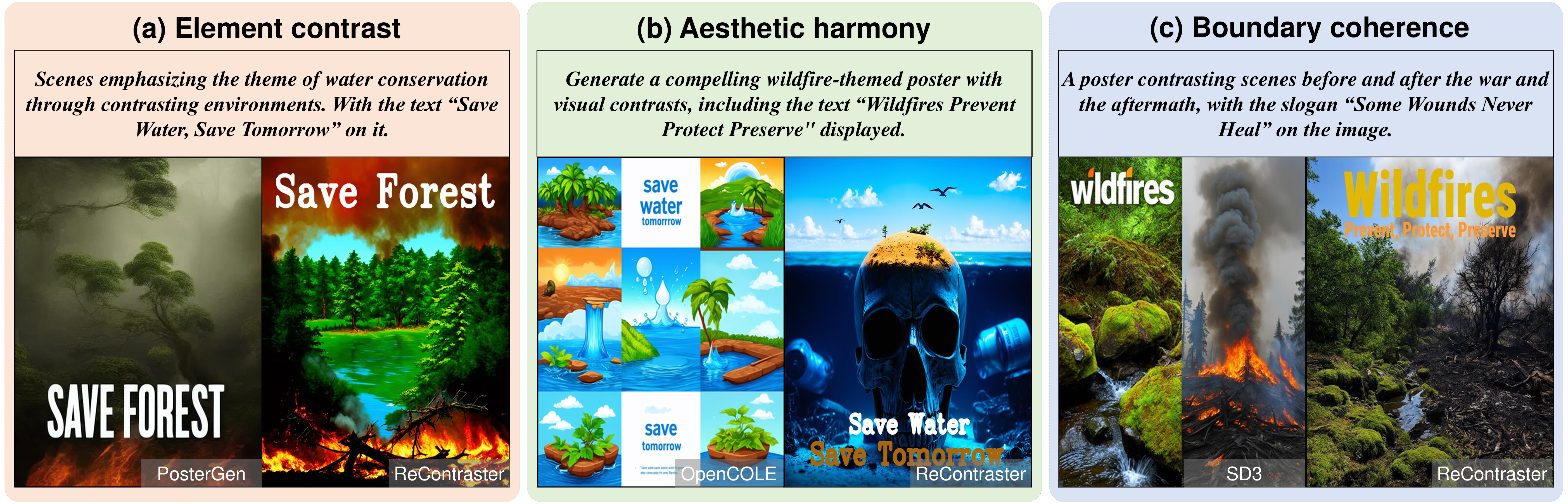}
  \caption{Illustration of our ReContraster for poster generation. Given a text description of the theme and visual texts, and a mask indicating region divisions, our model produces:   (a) Visually striking poster designs with highly contrastive elements, compared to PosterGen. (b) Structurally balanced compositions with aesthetic harmony, compared to OpenCOLE. (c) Seamless and coherent transitions across region boundaries, compared to SD3.
  }
  \label{fig:teaser}
\end{figure*}

Applying the ``contrast effect'' is a promising approach for creating striking posters. However, existing methods~\cite{SD3, opencole, posterllava} struggle to reason about contrastive elements from the user-provided theme while preserving visual harmony, often producing disjointed results.
In this paper, we propose \textbf{ReContraster} (\textbf{Re}gional \textbf{Contrast} Post\textbf{er}) to address this problem. 
As illustrated in Fig.~\ref{fig:teaser}, ReContraster has the following primary properties:
\textit{(i)} \textbf{Element contrast:} 
ReContraster crafts visually striking results by designing contrastive elements and colors (Fig.~\ref{fig:teaser}~(a), contrast between ancient towns and modern cities).
\textit{(ii)} \textbf{Aesthetic harmony:} 
ReContraster produces aesthetically appealing posters with well-balanced and organized layouts (Fig.~\ref{fig:teaser}~(b), shared silhouette between a skull and an island).
\textit{(iii)} \textbf{Boundary coherence:}
ReContraster synthesizes posters with smooth and artifact-free regional transitions based on user-specified region divisions (Fig.~\ref{fig:teaser}~(c), transition from a greenery landscape to a burned one).

To the best of our knowledge, ReContraster is the first training-free model capable of generating posters with all the properties above. 
By emulating the cognitive behaviors of a poster designer, ReContraster introduces the \textbf{compositional multi-agent system}, which comprises three specialized agents:
\textit{(i)} A cognition agent establishes elemental contrast by reasoning about the theme to identify contrastive element pairs;
\textit{(ii)} An arranger agent organizes the layout to ensure aesthetic harmony among the selected contrastive element pairs;
and \textit{(iii)} A refiner agent iteratively optimizes the poster, aiming to strike a balance between compelling visual contrast and overall compositional harmony.
To further address the challenge of boundary coherence, we introduce the \textbf{hybrid denoising strategy}. During the initial denoising steps, a gradient consistency loss is applied to penalize gradient differences across boundaries. 
Subsequently, we employ a joint regional denoising that linearly blends noise from adjacent regions to guarantee harmonious transitions.

For comprehensively evaluating ReContraster, we contribute a new benchmark dataset containing diverse images, each with a corresponding text description and a region division annotation. We conduct four human evaluation experiments to ensure element contrast, content continuity, division accuracy, and text-image consistency. Our contribution could be summarized as follows: 

\begin{itemize}
    \item We propose ReContraster, the first training-free model to leverage regional contrast to make posters stand out. Extensive experiments demonstrate superior performance.

    \item We develop compositional multi-agent system that constructs three agents to separately identify elements, organize layout, and evaluate generated poster candidates.
    
    \item We design hybrid regional denoising to maintain boundary coherence across region boundaries. A benchmark dataset is contributed for comprehensive evaluation.
\end{itemize}

\section{Related Work}

\subsection{Poster Generation}
Poster generation aims to craft visual impact with message clarity.
Existing methods can be divided into two main categories: content-aware and complete poster generation.
On the one hand, content-aware methods~\cite{cglgan, contentgan, dsgan} focus on arranging elements on a given background. This approach has been recently advanced by LLM-based models (\eg, LayoutPrompter~\cite{layoutprompter} and PosterLLaVa~\cite{posterllava}), which provide a more nuanced semantic understanding.
However, the reliance on a pre-selected background fundamentally limits their creative flexibility and application scenarios.
On the other hand, complete poster generation methods offer a more versatile solution, creating posters solely from text prompts (\eg, PosterGen~\cite{posterllava}, COLE~\cite{cole}, and OpenCOLE~\cite{opencole}). Although this removes the constraint of a background image, they often struggle to generate aesthetically compelling posters.
To overcome these challenges, our work focuses on complete poster generation, with the additional objective of enhancing visual impact through regional contrast.

\subsection{Large Language Model Agent}
Large Language Models (LLMs) have evolved into powerful agents capable of leveraging reasoning and planning abilities for a wide range of tasks, including basic human activities (\eg, searching~\cite{yao2023react,qiao2024autoact,cai2022query}, communication~\cite{xu2024rethinking,deng2024multi}, and visual processing~\cite{gpt4,llava,cheng2024seeclick}) and more complex skills (\eg, logical and mathematical reasoning~\cite{wei2022chain}, task decomposition~\cite{zhang2023exploring,stage}, and tool utilization~\cite{genartist}).
With these advancements, LLM agents are introduced into poster design for the generation of visual texts (\eg, COLE~\cite{cole} and OpenCOLE~\cite{opencole}) and poster layout (\eg, LayoutPrompter~\cite{layoutprompter}, PosterLlama~\cite{posterllama}, PosterLLaVa~\cite{posterllava}, and postermaker~\cite{postermaker}).
However, existing methods are primarily single-agent systems that struggle with the multifaceted nature of poster design, particularly in balancing visually striking elements with aesthetically appealing compositions. To address this limitation, we propose a multi-agent system to collaboratively identify contrastive elements, organize poster layout, and achieve a balance between regional contrast and overall harmony.

\subsection{Text-driven Image Generation}
Text-driven image generation has become a central focus of recent visual content synthesis~\cite{ding2021cogview, ramesh2021zero, aif}.
Recent progress has been significantly accelerated by diffusion models, with notable examples like Stable Diffusion~\cite{stablediffusion, sdxl, chen2023pixart, deer, aiedit} generating high-quality images by iteratively denoising Gaussian noise conditioned on textual prompts. 
TextDiffuser-2~\cite{textdiffuser2} and GlyphDraw~\cite{glyphdraw2} enhance visual text clarity, while AnyText~\cite{anytext} and UDiffText~\cite{udifftext} progressively expand support to multiple languages. 
To further improve the controllability, researchers explore the use of adapters to tailor diffusion models for specific tasks (\eg, attribute control~\cite{t2i-adapter, ip-adapter}, style control~\cite{style-adapter}, and spatial control~\cite{creatilayout}). Concurrently, alternative approaches~\cite{lcad, voynov2023sketch} focus on refining the sampling strategy to improve image quality and consistency.
To generate regional contrast posters with boundary coherence, we follow prior research to propose the hybrid regional denoising, a novel sampling inference to produce visually harmonious transitions across region boundaries.

\begin{figure*}[t]
   \centering
  \includegraphics[width=\linewidth]{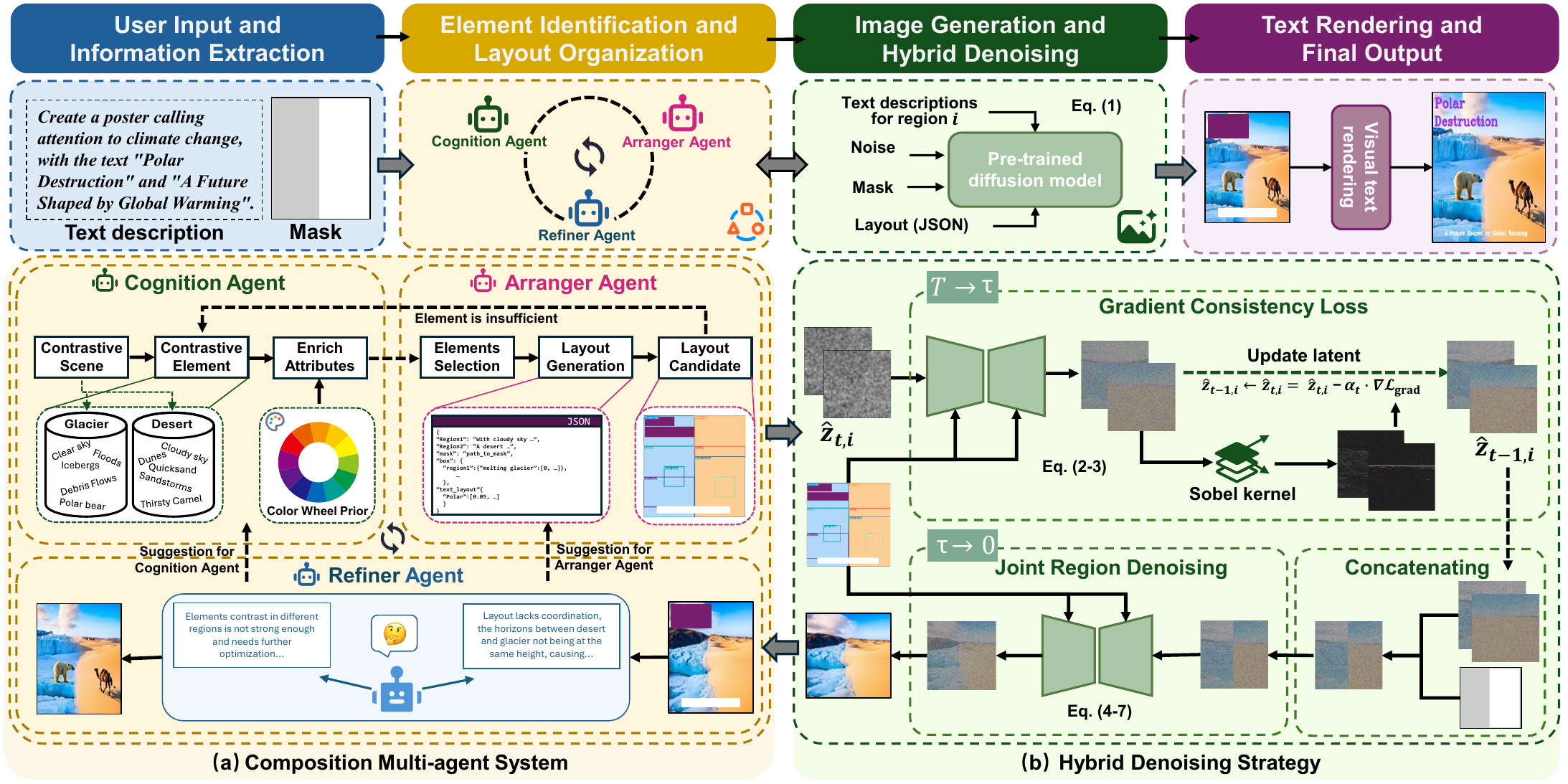}
  \caption{
  Given a text description and a mask indicating region divisions, ReContraster initially uses an LLM~\cite{gpt4} to extract the theme and visual texts separately.
  The extracted theme is then processed by the \textbf{compositional multi-agent system} in an iterative refinement loop. Within each iteration, the cognition agent identifies contrastive elements, the arranger agent organizes a layout in a JSON format, and a candidate image is generated based on this layout.
  This candidate is then evaluated by the refiner agent, which provides feedback to guide subsequent rounds of improvement until a design is approved.
  The generation of these candidate images relies on a pre-trained diffusion model equipped with the \textbf{hybrid denoising strategy}, where the gradient consistency loss and joint region denoising are applied to maintain content continuity and ensure harmonious transitions across region boundaries, respectively.
  The visual texts are finally rendered onto the generated image to produce a visually striking and aesthetically appealing poster.
  }
  \vspace{-2mm}
  \label{fig:pipeline}
\end{figure*}

\section{Methodology}

\subsection{Overview} \label{sec:overview}

As shown in Fig~\ref{fig:pipeline}, ReContraster is systematically organized into the following stages:


\noindent \textit{(i)} \textbf{User input and information extraction.} Initially, users provide a text description and a corresponding mask indicating region divisions. We use an LLM~\cite{gpt4} to extract the theme and visual texts separately.

\noindent \textit{(ii)} \textbf{Element identification and layout organization.}
This stage is handled by our compositional multi-agent system through an iterative process. Specifically, the extracted theme is passed to the cognition agent, which identifies potential contrastive element pairs. Subsequently, the arranger agent selects appropriate element pairs and organizes them into a conceptual poster layout. This proposed layout then guides an initial image generation in \textit{Stage (iii)}. The resulting image is fed back to the refiner agent to evaluate its overall contrast and harmony. If the result is unsatisfactory, the refiner provides a feedback to the cognition and arranger agents to generate an improved results. This loop continues until the refiner agent approves the composition.

\noindent \textit{(iii)} \textbf{Image generation and hybrid denoising.}
Guided by the approved layout from \textit{Stage (ii)}, a pre-trained diffusion model is used to concrete the selected visual elements (Eq.(\ref{eq:firstdenoise})). To ensure seamless regional transitions, we then apply our hybrid denoising strategy. This involves using a gradient consistency loss (Eqs.~(\ref{eq:gradloss}-\ref{eq:updatelatent})) for content continuity and a joint regional denoising technique for harmonious transitions (Eqs.~(\ref{eq:latentfuse}–\ref{eq:seconddenoising})).

\noindent \textit{(iv)} \textbf{Text rendering and final output.}
In the final stage, the visual texts specified in \textit{Stage (i)} are rendered onto the generated image by adopting the text rendering method (\ie, OpenCole~\cite{opencole}), producing a visually striking and aesthetically appealing poster.

\subsection{Compositional Multi-agent System}
To emulate the cognitive behaviors of a poster designer, we propose the compositional multi-agent system, as shown in Fig.~\ref{fig:pipeline}~(a). Specifically, we introduce three specialized agents: cognition agent, arranger agent, and refiner agent, which collaborate in an iterative loop to design the poster's visual content and layout.

\noindent \textbf{Cognition agent.}
The cognition agent is responsible for transforming the extracted theme into a set of visually contrastive elements. 
It utilizes an LLM~\cite{gpt4} through a Chain-of-Thought (CoT) \cite{kojima2022large,li2024enhancing} process. 
Specifically, for a given theme $T$ (\eg, calling attention to climate change), the LLM first identifies a pair of high-level contrastive scenes (\eg, glacier and desert) to establish the core thematic conflict.
Based on these scenes, the LLM then brainstorms specific objects and forms initial contrastive element pairs (\eg, sandstorm and icebergs).
Finally, the LLM enriches these element pairs with detailed color attributes, drawing upon principles from the color wheel~\cite{cohen2006color} are incorporated as prior knowledge. The color wheel defines color relationships by position, distinguishing between complementary colors (\eg, orange and blue) and analogous colors (\eg, purple and blue). The LLM is instructed to apply these principles by assigning complementary colors to the core elements to maximize their visual opposition (\eg, a yellow sandstorm and a blue iceberg), while leveraging analogous colors to ensure harmonious transitions at region boundaries.

\noindent \textbf{Arranger agent.}
The arranger agent selects optimal element pairs from the candidates provided by the cognition agent and organizes them into an aesthetically harmonious composition. The final layout is structured as a JSON format, generated by simultaneously optimizing for several critical factors, including \textit{shape harmony} by considering element shapes for smooth boundary transitions; \textit{style unification} by ensuring consistent visual styles, textures, and color schemes; \textit{semantic clustering} by grouping related elements for narrative logic; and \textit{typographic integration} by allocating sufficient space for effective text placement. 
Following these principles, the arranger agent produces an aesthetically harmonious layout to guide the image generation process. 
Furthermore, when the initial elements prove insufficient to meet these compositional criteria, the agent can issue a targeted request to the cognition agent for additional element pairs.

\noindent \textbf{Refiner agent.}
The refiner agent serves as a critical evaluator, guiding the iterative refinement of the generated image. Upon receiving an image, the agent assesses two key aspects:
\textit{Visual contrast.} The agent evaluates the clarity and distinctness of elements. If it detects insufficient contrast, it instructs the cognition agent to regenerate element pairs with greater visual differentiation.
\textit{Aesthetic harmony.} The agent examines the overall composition for aesthetic imperfections. If improper alignment or imbalanced visual weight are identified, it provides the arranger agent with corrective guidance for optimization.
This iterative feedback loop continues until the image satisfies all predefined quality thresholds or a maximum number of iterations is reached.

\subsection{Hybrid Denoising Strategy} \label{sec:coherence}

To concrete contrastive elements in their corresponding regions within the poster, we propose the two-stage hybrid denoising strategy, as illustrated in Fig.~\ref{fig:pipeline}~(b). Given a pre-trained diffusion model, the first stage focuses on concreting visual content based on the selected and arranged elements, while a gradient consistency loss ensures content continuity across region boundaries. The second stage performs joint denoising across all regions, achieving harmonious transitions and overall visual coherence for the poster.

\noindent \textbf{Gradient consistency loss.} 
To ensure that visual content corresponds accurately to the contrastive elements, we independently concrete visual content during the initial $\tau$ denoising steps. Given the latent code $\hat{z}_{t,i}$ at denoising step $t$ and the corresponding text descriptions $\mathcal{P}_i$ of the visual content at region $i$, the denoising process is formulated as:
\begin{equation}
    \footnotesize \hat{z}_{t-\Delta t, i} = \hat{z}_{t, i} - \Delta t \cdot v_\theta(\hat{z}_{t, i}, t, \mathcal{P}_{i}),
\label{eq:firstdenoise}
\end{equation}
where $v_\theta$ represents the pre-trained velocity prediction model~\cite{SD3} parameterized by $\theta$, $t$ denotes the continuous time in the ODE framework, and $\Delta t$ is the integration step size for the solver.

However, simply concatenating separately denoised latent codes often leads to visible discontinuities at region boundaries. Therefore, we introduce a gradient consistency loss applied after each denoising step to enforce harmonious transitions by penalizing gradient differences across these boundaries. Specifically, we calculate gradient matrices using a $5 \times 5$ Sobel kernel~\cite{sobel} along boundaries according to region divisions. The gradient consistency loss is defined as:
\begin{equation}
    \footnotesize \mathcal{L}_{\textrm{grad}} = \sum_{b \in \mathcal{B}} \Big( 1 - \textrm{Avg} \big(\textrm{Cos}(\mathcal{G}_{t,b}, \mathcal{G}_{t,b}^\prime)^2 \big) \Big), \label{eq:gradloss}
\end{equation}
where $\mathcal{B}$ is the set of boundary indices, $\mathcal{G}_{t,b}$ and $\mathcal{G}_{t,b}^\prime$ denote the calculated gradient matrices at denoising step $t$ for the regions along boundary index $b$, $\textrm{Cos}$ represents the cosine similarity, and $\textrm{Avg}$ denotes the averaging operator. This loss encourages gradient alignment across boundaries. We update each latent code as follows:
\begin{equation}
    \footnotesize \hat{z}_{t,i} = \hat{z}_{t,i} - \alpha_{t} \cdot \nabla \mathcal{L}_{\mathrm{grad}}. \label{eq:updatelatent}
\end{equation}
After the initial $\tau$ steps, the latent codes are concatenated based on the region divisions. We empirically set $\tau=10$.

\noindent \textbf{Joint region denoising.} 
While the gradient consistency loss ensures content continuity, it does not guarantee harmonious transitions and visual coherence. Considering that the first stage (the initial $\tau$ steps) provides a coarse layout of the visual elements, the second stage focuses on the overall harmonization of the different regions to improve the harmonization of the generated poster. 

Denoting the binary mask for the region $i$ as $\mathcal{M}_i$, we first concatenate the denoised latent codes at denoising step $\tau$ based on the region divisions:
\begin{equation}
    \footnotesize z_{\tau} =  \sum_{i \in \mathcal{V}} (\hat{z}_{\tau,i} \odot \mathcal{M}_{i}), \label{eq:latentfuse}
\end{equation}
where $\mathcal{V}$ is the set of regions and $\odot$ is element-wise multiplication.
Subsequently, we require the visual content in regions near the boundaries to blend smoothly. Let $r$ represent the boundary margin and $d_{i,j}$ represent the shortest distance from each point in region $i$ to the boundary of the adjacent region $j$, we calculate the distance-weighted noise predictions:
\begin{equation}
    \footnotesize \!\!\! \hat{\epsilon}_{i, j} = \dfrac{r+d_{i,j}}{2r} \cdot v_\theta(z_t, t, \mathcal{P}_i) + \dfrac{r-d_{i,j}}{2r} \cdot v_\theta(z_t, t, \mathcal{P}_j), \!\! \label{eq:mix}
\end{equation}
where $\mathcal{P}_i$ and $\mathcal{P}_j$ are the text descriptions for region $i$ and $j$, respectively, and $d_{i,j}$ is clipped to the range $[0, r]$. 
Next, we incorporate the influence of all adjacent regions following the classifier-free guidance calculation~\cite{classifier-free}:
\begin{equation}
    \footnotesize \!\!\! \hat{\epsilon}_i  =  v_\theta(z_t, t, \emptyset)  +  \frac{w}{|\mathcal{A}(i)|}  \cdot   \sum_{j \in \mathcal{A}(i)} \big ( \hat{\epsilon}_{i, j}  -  v_\theta(z_t, t, \emptyset) \big), \!\! \label{eq:multimix}
\end{equation}
where $\mathcal{A}(i)$ is the set of regions adjacent to region $i$.
Finally, we reformulate the denoising process as:
\begin{equation}
    \footnotesize \!\!\! {z}_{t-1} = \frac{1}{\sqrt{\alpha_{t}}} 
    \big ({z}_{t} - \frac{1-\alpha_{t}}{\sqrt{1-\bar{\alpha}_{t}}} \cdot 
    \sum_{i \in \mathcal{V}} (\hat{\epsilon}_i \odot \mathcal{M}_i )  \big ) + \sigma_{t}\epsilon. \!\! \label{eq:seconddenoising}
\end{equation}
In practice, we set $r=1/32$ of the latent space 
size and $w=3$ according to experimental experience.

\section{Benchmark Dataset} \label{sec:dataset}

Existing datasets for poster generation provide annotations for poster structure layout~\cite{posterllava, cglgan}, visual text arrangements~\cite{autoposter}, and theme descriptions~\cite{mpds_dataset}.
However, they primarily focus on movie posters and human figures, generally lacking regional contrast, a critical feature for crafting visually striking posters. 
To address this limitation, we collect a high-quality benchmark dataset tailored for regional contrast posters.
Specifically, we carefully select samples from existing datasets~\cite{posterllava, cglgan, autoposter, mpds_dataset}, e-commerce platforms, and Instagram, ensuring all posters are used in accordance with their respective terms of service and licensing agreements. We utilize GPT-4o~\cite{gpt4} and SAM~\cite{segmentanything} to annotate text descriptions and region divisions for each poster, followed by manual verification and refinement. This process results in a dataset of 643 posters for model evaluation\footnote{The benchmark dataset will be released upon publication, with representative samples visualized in Supp.}. 


We conduct four human evaluation experiments to assess the quality of our benchmark dataset, focusing on whether: 
the collected posters contain sufficient element contrast (Exp-I); 
the collected posters demonstrate the overall aesthetic harmony (Exp-II); 
the annotated text descriptions accurately describe the theme of each poster (Exp-III); and
the annotated region divisions accurately reflect the structure of each poster (Exp-IV).
For each experiment, we randomly select 100 samples from our dataset and ask 25 volunteers to independently evaluate these samples, assigning each a rating of ``Failed'', ``Borderline'', ``Acceptable'', or
``Perfect''. As shown in Tab.~\ref{tab:data_evaluation}, over 90\% of volunteers rate the annotation quality as "Acceptable" or higher across all four aspects, confirming the dataset quality.

\section{Experiments}
\subsection{Experimental Setup}
Since ReContraster is a training-free model\footnote{The analysis of model efficiency is detailed in Supp.}, it is capable of seamlessly integrating advanced models for its pipeline.
In our experiments, we use GPT-4o~\cite{gpt4} as the large language model, CreatiLayout~\cite{creatilayout} for image generation, and OpenCOLE~\cite{opencole} for text rendering\footnote{The workflow and prompts are provided in Supp.}.
Notably, since related models are not designed for regional contrast, we prompt the GPT-4o~\cite{gpt4} to directly describe the selected elements in each region and their corresponding visual effects. This setup potentially favors the comparison methods.

\begin{table}[t]
\caption{Percentage (\%) of user ratings in the four experiments of human evaluation for the collected dataset. }
\vspace{-6mm}
\begin{center}
{
    \setlength\tabcolsep{12pt}
    \centering
    \begin{adjustbox}{width={0.48\textwidth},totalheight={\textheight},keepaspectratio}
    \begin{tabular}{lcccc} \cr \toprule
    Rating & Exp-I & Exp-II & Exp-III & Exp-IV  \\ \midrule
        Failed & $1.28$ & $0.44$ & $0.00$ & $0.84$   \\ 
        Borderline & $6.92$ & $6.44$ & $5.68$ & $2.60$   \\ 
        Acceptable & $36.56$ & $29.64$ & $25.72$ & $10.92$   \\ 
        Perfect & $\mathbf{55.24}$ & $\mathbf{63.48}$ & $\mathbf{68.60}$ & $\mathbf{85.64}$   \\ 
        \bottomrule
\end{tabular}\label{tab:data_evaluation}
    \end{adjustbox}
}
\end{center}
\vspace{-4mm}
\end{table}

\begin{figure*}[t]
   \centering
  \includegraphics[width=\linewidth]{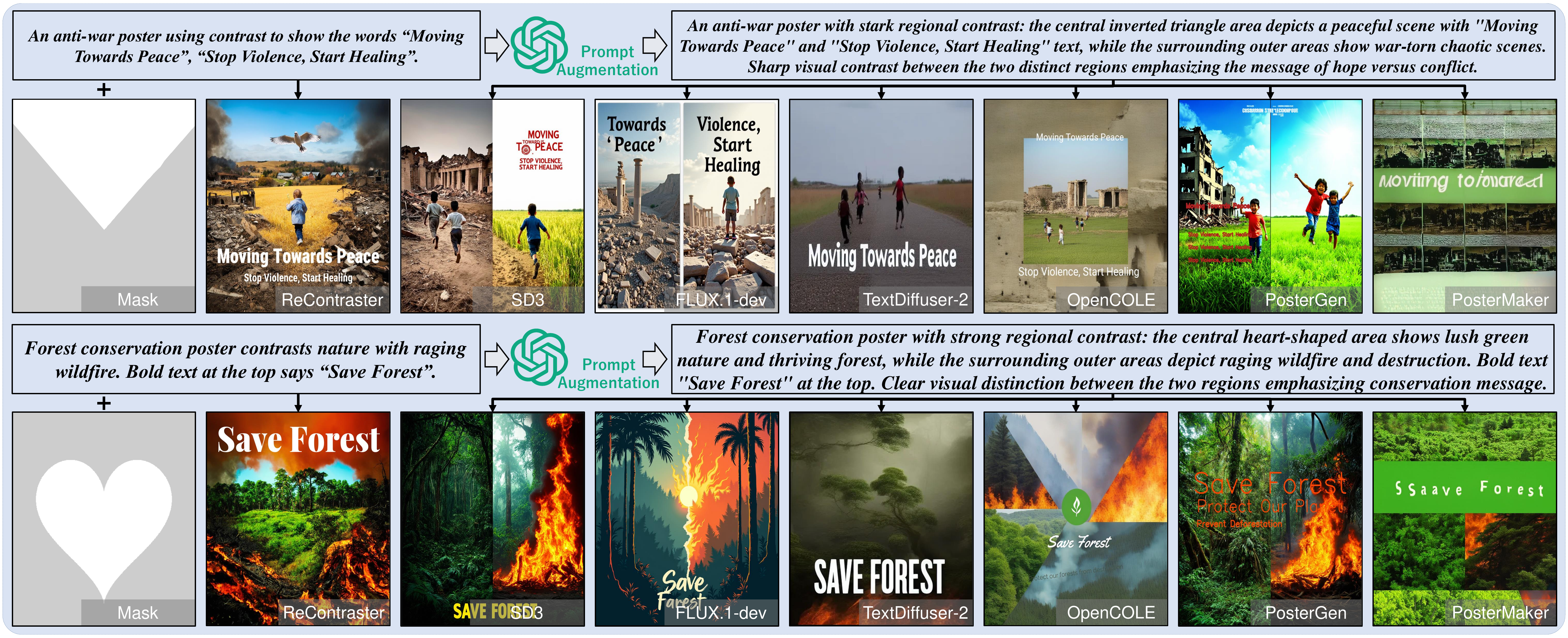}
  \caption{Visual quality comparisons with text-to-image generation methods and poster generation methods.}
  \label{fig:comparison}
  \vspace{-2mm}
\end{figure*}

\subsection{Quantitative Evaluation Metrics} \label{sec:Quantitative_Evaluation_Metrics}
We comprehensively evaluate ReContraster across following aspects: 
\textit{(i)} 
Following LAION-5B~\cite{laion}, we calculate the \textbf{LAlON Aesthetic Score (LAS)} to measure the aesthetic quality of generated posters.
\textit{(ii)} 
Following StyTr2~\cite{stytr2}, we compute the \textbf{Regional Style Difference (RSD)} by extracting style features across regions using the VGG model~\cite{vgg} and calculate the MSE, measuring the visual contrast.
\textit{(iii)}
Using the Sobel kernel~\cite{sobel}, we calculate the \textbf{Boundary Gradient Difference (BGD)} by calculating the cosine similarity between gradients of patches along the region boundaries, measuring the content continuity.
\textit{(iv)} 
Following TextDiffuser~\cite{textdiffuser}, we recognize visual texts on the poster and compute the \textbf{Optical Character Recognition (OCR)} accuracy compared to the provided text, measuring the accuracy of text rendering. 
\textit{(v)}
To further evaluate the holistic quality of the generated posters, we employ the VLM~\cite{gpt4} to rate \textbf{Content Relevance and Effectiveness (CRE)}, \textbf{Visual Appeal and Impact (VAI)}, and \textbf{Boundary Integration and Harmony (BIH)} on a scale of $1$ to $10$ score\footnote{Details are provided in Supp.}

\begin{table*}[t]
\caption{Quantitative experiment results of comparison, ablation, and user study. $\uparrow$ ($\downarrow$) means higher (lower) is better. The best performances are highlighted in \textbf{bold}.}
\vspace{-3mm}
\begin{center}
{
    \setlength\tabcolsep{4pt}
    \centering
    \begin{adjustbox}{width={0.85\textwidth},totalheight={\textheight},keepaspectratio}
    \begin{tabular}{l | c c c c  c c c | c c c c} \toprule
    \multirow{2}{*}{Method} 
    & \multicolumn{7}{c|}{\textbf{Quantitative Evaluation Metrics}} & \multicolumn{4}{c}{\textbf{User Study}} \cr
    & LAS $\uparrow$ & RSD $\uparrow$ & BGD $\downarrow$ & OCR $\uparrow$ & CRE $\uparrow$ & VAI $\uparrow$ & BIH $\uparrow$ & Exp-I & Exp-II & Exp-III & Exp-IV \cr \midrule
    SD3 & $4.2810$ & $596.69$ & $0.0566$ & $0.55$ &  $7.61$ &  $7.42$ &  $6.30$ &  $11.76$ & $14.80$ & $15.16$ & $2.04$\cr 
    Flux.1-dev & $4.8938$ & $602.49$ & $0.0460$ & $0.58$ & $7.41$ &  $7.64$ &  $6.62$ &  $12.36$ & $16.48$ & $12.24$ & $3.08$\cr
    TextDiffuser-2 & $4.3693$ & $446.47$ & $0.0818$ & $0.56$ & $5.30$ &  $6.76$ &  $4.85$ & $5.04$ & $5.32$ & $3.12$ & $0.92$\cr 
    OpenCOLE & $3.6768$ & $286.74$ & $0.0548$ & $0.57$ & $7.45$ &  $5.82$ &  $6.33$ &  $3.28$ & $7.24$ & $14.20$ & $1.48$\cr 
    PosterGen & $3.4895$ & $717.80$ & $0.0674$ & $0.43$ & $7.52$ &  $5.11$ &  $3.17$ &  $14.12$ & $10.84$ & $13.08$ & $1.92$\cr 
    PosterMaker & $4.5686$ & $600.94$ & $0.0759$ & $0.53$ & $7.02$ &  $6.69$ &  $4.26$ &  $12.08$ & $5.68$ & $10.96 $ & $1.20$\cr
    ReContraster & $\mathbf{5.0966}$ & $\mathbf{842.60}$ & $\mathbf{0.0375}$ & $\mathbf{0.65}$ & $\mathbf{7.82}$ & $\mathbf{7.87}$ & $\mathbf{7.04}$ & $\mathbf{41.36}$ & $\mathbf{39.64}$ & $\mathbf{31.24}$ & $\mathbf{89.36}$ \\ \midrule

    \textit{W/o} CAI & $4.7501$ & $766.02$ & $0.0388$ & $0.56$ & $7.04$ &  $7.64$ &  $6.94$ &  $N/A$ &  $N/A$ &  $N/A$ &  $N/A$\cr
    \textit{W/o} IRA & $4.7280$ & $676.95$ & $0.0446$ & $0.57$ & $7.33$ &  $7.57$ &  $6.73$ &  $N/A$ &  $N/A$ &  $N/A$ &  $N/A$\cr
    \textit{W/o} GCL & $4.9363$ & $789.86$ & $0.0482$ & $0.60$ & $7.38$ &  $7.71$ &  $6.64$ &  $N/A$ &  $N/A$ &  $N/A$ &  $N/A$\cr 
    \textit{W/o} JRD & $4.8387$ & $804.52$ & $0.0467$ & $0.62$ & $7.51$ &  $7.68$ &  $6.58$ &  $N/A$ &  $N/A$ &  $N/A$ &  $N/A$\cr
    \bottomrule
    \end{tabular}\label{tab:comparison}
    \end{adjustbox}
    \vspace{-2mm}
}
\end{center}
\end{table*}

\begin{figure*}[t]
  \centering
  \includegraphics[width=0.92\linewidth]{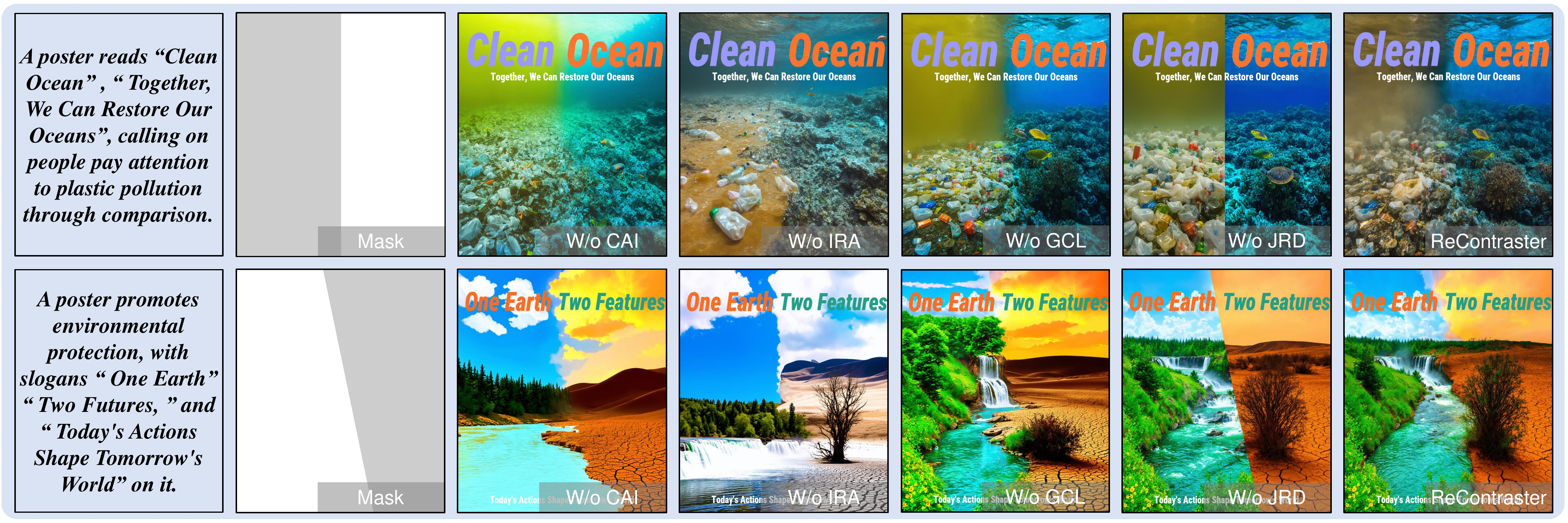}
  \caption{Ablation study results with different variants of ReContraster. }
  \label{fig:ablation}
  \vspace{-2mm}
\end{figure*}

\subsection{Comparison with Relevant Methods}
As the first approach to regional contrast poster generation, we conduct comparison experiments with related text-to-image generation methods (\ie, SD3 \cite{SD3}, Flux.1-dev \cite{flux}, and TextDiffuser-2 \cite{textdiffuser2}) and complete poster generation methods (\ie, OpenCOLE \cite{opencole}, PosterGen \cite{posterllava}, and PosterMaker \cite{postermaker}).

\noindent \textbf{Qualitative comparisons.} 
In Fig.~\ref{fig:comparison}, we present visual quality comparisons with several existing methods. 
SD3 fails to create a harmonious transition across the two contrastive regions, exhibiting a discontinuity in content (second row, an abrupt division between vibrant greenery and parched land).
Flux.1-dev suffers from insufficient element and color contrast between the two regions, resulting in a visually indistinct composition (first row, both regions are dominated by similar grayish-brown hues). 
TextDiffuser-2 struggles to generate semantically consistent elements with text descriptions (second row, the obscured though well-substantial tree and blazing fire undermine the theme of ``Save Forest'').
OpenCOLE demonstrates suboptimal element arrangement and a compromised compositional structure (first row, children playing in a field incongruously are surrounded by a dull and blurred background).
PosterGen struggles with visual text legibility since the colors of the visual text and the background are not easily distinguishable (second row, indistinct red visual texts are rendered on a fire background).
PosterMaker generates chaotic and semantically incoherent content, damaging the overall aesthetic harmony (second row, the forest and fire elements are placed in a fragmented layout).
In contrast, our results effectively balance element contrast and aesthetic harmony, while guaranteeing boundary coherence, producing visually striking and aesthetically appealing posters.

\noindent \textbf{Quantitative comparisons.} 
We present quantitative comparisons in Tab.~\ref{tab:comparison}, demonstrating that our method outperforms all compared methods across all metrics. ReContraster achieves highly aesthetically appealing results (LAS), exhibits visually striking element contrasts (RSD) with harmonious transitions across region boundaries (BGD), and accurately renders the visual texts (OCR). Additionally, ReContraster achieves top scores across all three VLM-based scores (CRE, VAI, and BIH).

\noindent \textbf{User study.}
We conduct four user studies to evaluate whether ReContraster is preferred over state-of-the-art methods, using the same evaluation criteria introduced for the benchmark dataset.
For each experiment, participants are shown a text description and seven generated posters. 
We conduct these experiments on Amazon Mechanical Turk (AMT), randomly selecting 100 samples from the benchmark dataset. Experiment results are polled by 25 volunteers independently. We present these scores in Tab.~\ref{tab:comparison}, highlighting the subjective advantages of our approach.

\subsection{Ablation Study}
We remove various proposed modules and designed losses to construct four baselines to evaluate their individual contributions. The evaluation scores and generated posters are shown in Tab.~\ref{tab:comparison} and Fig.~\ref{fig:ablation}.

\noindent \textbf{W/o Cognition \& Arranger Interaction (CAI).} 
We replace the cognition and arranger agents with a single LLM to eliminate interaction between these agents. As shown in Fig.~\ref{fig:ablation} second row, this baseline generates posters with less compelling elements, resulting in a lower LAS score.

\noindent \textbf{W/o Iterative Refiner Agent (IRA).} 
We discard the refiner agent, thereby eliminating the iterative loop. As shown in Fig.~\ref{fig:ablation} first row, this baseline reduces the contrast effect between elements, leading to a lower RSD score.

\noindent \textbf{W/o Gradient Consistency Loss (GCL).} 
We discard the gradient consistency loss during inference. As shown in Fig.~\ref{fig:ablation} second row, this baseline results in visual discontinuities across the region boundary, leading to a worse BGD score. 

\noindent \textbf{W/o Joint Region Denoising (JRD).} 
We remove the joint region denoising during inference. As shown in Fig.~\ref{fig:ablation} first row, this baseline leads to disharmonious transitions across region boundaries, and reduces the LAS score.

\begin{figure}[t]
  \centering
  \includegraphics[width=\linewidth]{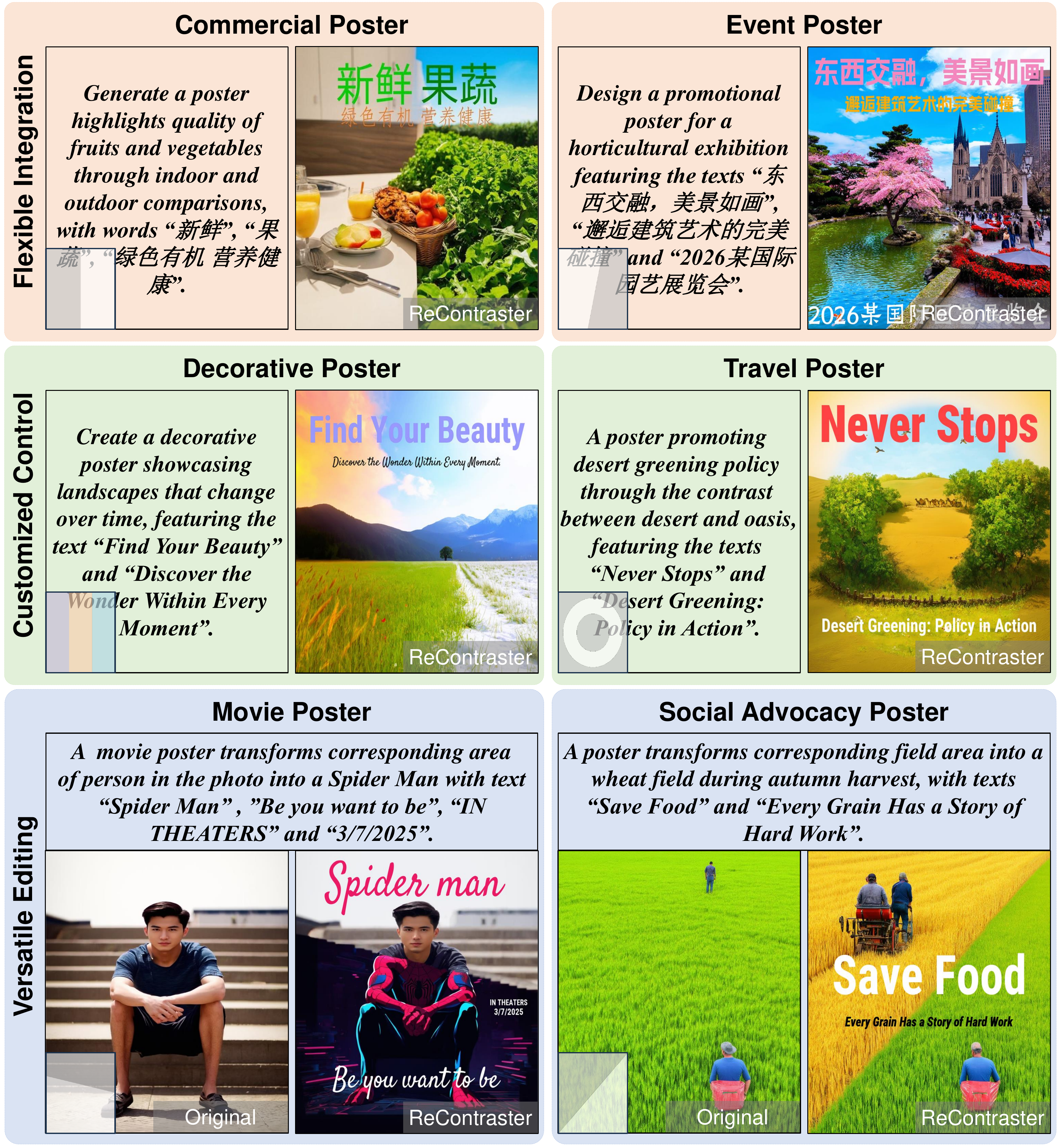}
  \caption{Application scenarios of ReContraster.}
  \label{fig:application}
  \vspace{-2mm}
\end{figure}

\subsection{Application} \label{sec:application}
As shown in Fig.~\ref{fig:application}, ReContraster supports flexible integration of multi-lingual text descriptions, offers customized control over region divisions, and provides versatile editing of user-provided images\footnote{Technical details are provided in Supp.}.

\section{Conclusion}
\label{sec:conclusion}
In this paper, we introduce ReContraster, a training-free model that leverages the principle of ``contrast effects'' to produce visually striking and aesthetically appealing posters. ReContraster achieves element contrast, aesthetic harmony, and boundary coherence via the proposed compositional multi-agent system and hybrid regional denoising modules. We contribute a new benchmark dataset and showcase application scenarios. Quantitative results confirm its superiority over relevant methods.

\section*{Limitations}
ReContraster performance is sensitive to user-specified region divisions, where overly small or highly complex divisions may diminish the visual appeal of the generated posters\footnote{Failure cases are presented in Supp.}. 
\bibliography{custom}

\end{document}